\documentclass[10pt,twocolumn,letterpaper]{article}
\usepackage{cvpr}
\usepackage{times}
\usepackage{epsfig}
\usepackage{graphicx}
\usepackage{amsmath}
\usepackage{amssymb}
\usepackage[numbers,sort&compress]{natbib}
\usepackage{mathrsfs}  
\usepackage{multirow}
\usepackage[ruled]{algorithm2e}
\LinesNumbered
\usepackage{booktabs}
\usepackage[breaklinks=true,bookmarks=false]{hyperref}
\hypersetup{colorlinks,linkcolor={blue},citecolor={blue},urlcolor={red}}

\cvprfinalcopy 

\newcommand{\cmtt}[1]{{\fontfamily{cmtt}\selectfont #1}}

\def\cvprPaperID{9} 

\begin{document}

\title{AutoCLINT: The Winning Method in AutoCV Challenge 2019}

\author{
Woonhyuk Baek\\
Kakao Brain\\
{\small \cmtt{wbaek@kakaobrain.com}}
\and
Ildoo Kim\\
Kakao Brain\\
{\small \cmtt{ildoo.kim@kakaobrain.com}}
\and
Sungwoong Kim\\
Kakao Brain\\
{\small \cmtt{swkim@kakaobrain.com}}
\and
Sungbin Lim\thanks{Corresponding Author}\\
UNIST\\
{\small \cmtt{sungbin@unist.ac.kr}}}

\maketitle

\begin{abstract}

NeurIPS 2019 AutoDL challenge is a series of six automated machine learning competitions. 
Particularly, AutoCV challenges mainly focused on classification tasks on visual domain. 
In this paper, we introduce the winning method in the competition, AutoCLINT (\textbf{C}omputationally \textbf{LI}ght \textbf{N}etwork \textbf{T}raining). The proposed method implements an autonomous training strategy, including efficient code optimization, and applies an automated data augmentation to achieve the fast adaptation of pretrained networks. 
We implement a light version of Fast AutoAugment \cite{lim2019fast} to search for data augmentation policies efficiently for the arbitrarily given image domains. 
We also empirically analyze the components of the proposed method and provide ablation studies focusing on AutoCV datasets. 
Our code\footnote{\url{https://github.com/kakaobrain/autoclint}} is open to the public, who wants to apply the proposed method.

\end{abstract}


\section{Introduction}

Due to the advance in computational resources and successes of machine learning in various domains, automated machine learning (AutoML) algorithms draw massive attention from the amounts of associations since they can reduce the exhaustive human efforts in ML engineering. 
Current AutoML researches mainly concentrate on hyperparameter tuning \cite{bergstra2011algorithms, hutter2019automated} and model search processes.
The desired AutoML system has to cover data exploration and data preprocessing before the training procedure without or less human intervention.

In this article, we introduce AutoCLINT, which won the first place in both Automated Computer Vision (AutoCV) competitions in NeurIPS 2019 AutoDL Challenge\footnote{\url{https://autodl.chalearn.org/}}.
The AutoDL challenge consists of a series of six challenges; AutoCV, AutoCV2, AutoNLP, AutoSpeech, AutoWeakly, and AutoDL.
All competitions are multi-label classification problems and are constrained by a time-budget and computational resource.
AutoCV and AutoCV2 challenges mainly focused on classification tasks on image and video data, respectively. 
Both competitions required an AutoML algorithm that can learn visual features universally under limited resources and constrained training time without human intervention. 
During the evaluation process, the participants could not identify nor directly observe the test image data except the image resolution and the number of samples.
These constraints consider the situation that users of the AutoML algorithms demand not only a ML model for arbitrary data but also a time-efficient training.
Additionally, a computationally cumbersome approach is impossible since only a single GPU is available during the evaluation process.
These restrictions are well-formulated by the AutoCV challenge design \cite{liu:hal-02265053}. See Appendix A for more details.
  
Inspired by the success of deep learning, neural architecture search (NAS) \cite{zoph2016neural, zoph2018learning, liu2018darts, kim2018scalable} is the most notable field in AutoML research. 
The recent NAS algorithm \cite{pham2018efficient} can find out neural architecture efficiently and outperforms a human-designed network in the image classification problem. 
However, searching optimal neural architecture with limited time and computational resources by NAS is still a challenging problem. 
In particular, NAS requires hundreds of GPU hours for extensive scale data, e.g., medical images such as CT or MRI \cite{kim2018scalable}.
An alternative approach, instead of NAS, is transfer learning. 
For computer vision tasks, ImageNet pre-trained models are widely used as the starting point. Notably, \cite{kornblith2019better} reports on the strength of ImageNet pretraining in image classification.    
Hence, AutoCLINT applied transfer learning since a fine-tuned model does not require training from scratch in the evaluation process. Empirical studies in Appendix show that this strategy was decisive in the competitions.

In addition to transfer learning, AutoCLINT applied a modified Fast AutoAugment \cite{lim2019fast} algorithm to find data augmentation policies automatically for a given image domain.
The original algorithm uses a distributed Bayesian optimization search, which requires a multi-GPU environment. 
Hence, we implemented a light-version of Fast AutoAugment, which is adapted for AutoCV competition. 
This algorithm makes it possible to find optimal data augmentation policies under a restricted time budget.
Notably, some image domains get better scores than no augmentation since these image domains have less number of samples. 

The main contribution of this paper is to provide the end-to-end training code for image classification by AutoML, which is easily transferrable under restricted time budget and computation resources. 
Our code includes a light version of Fast AutoAugment \cite{lim2019fast}.
We provide empirical studies in Appendix to validate the effect of each component of the proposed method. We believe this information can help non-ML experts to apply the proposed training method for their datasets.

\section{AutoCV Competition}
\label{sec:autocv_comp}

This section briefly introduces the description and evaluation metrics of AutoCV competition, the first series of NeurIPS 2019 AutoDL challenge.  
The goal of AutoCV competition is to find an automated solution that can handle any type of input image and generates a model that solves the associated task. 
Participants were asked to develop automatic methods for learning from raw visual information. 
We refer to the white paper of the AutoCV challenge design \cite{liu:hal-02265053} for more details.

\subsection{Competition Description}

\begin{figure}
     \centering
     \includegraphics[width=0.4\textwidth]{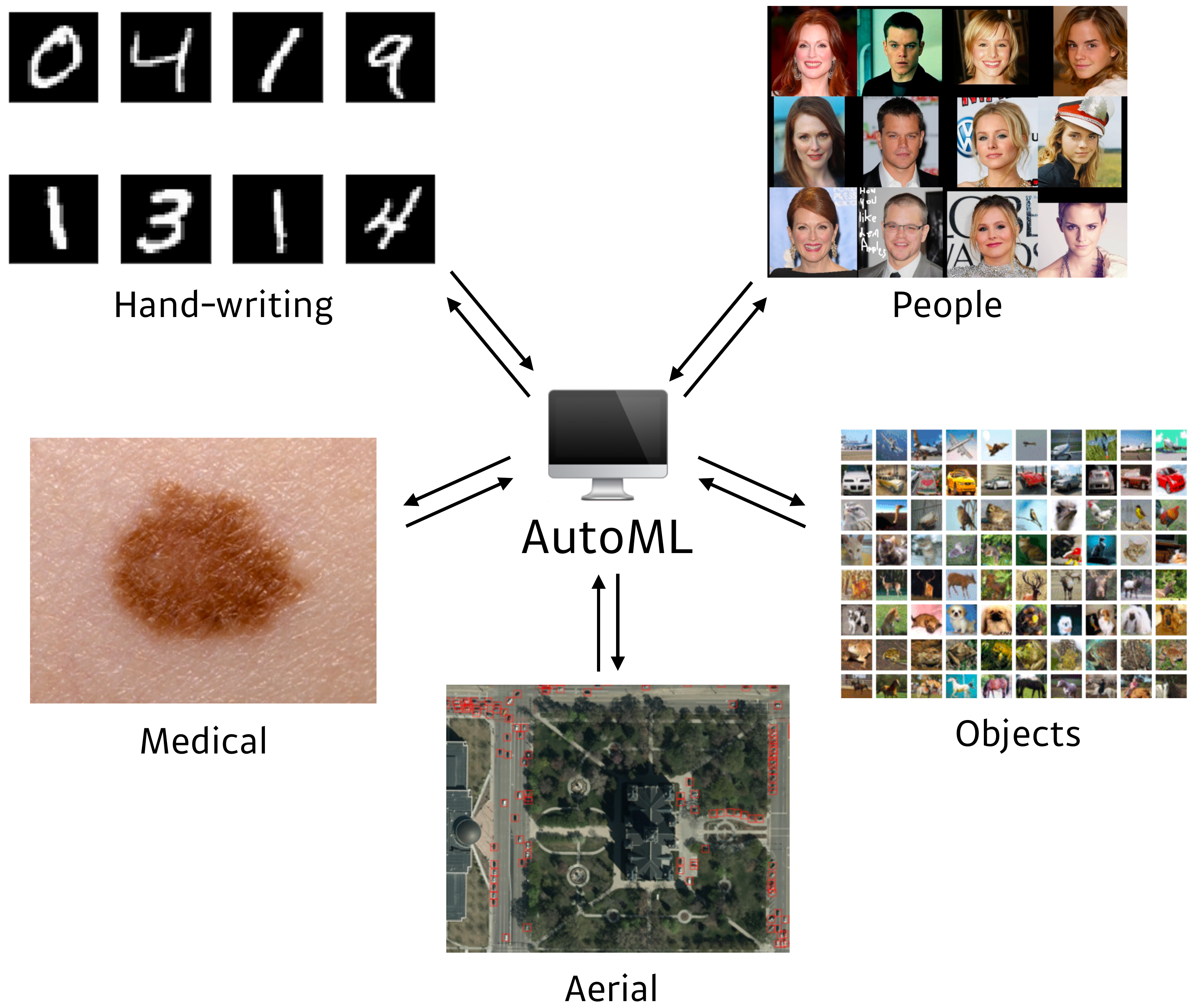}
     \caption{
     Examples of datasets in AutoCV challenge. Each participant submits a learning algorithm that can automatically conduct overall machine learning processes.  
     }
     \label{fig:dataset}
 \end{figure}

As previously mentioned, AutoCV competition requires universal learning machines, especially for image data, considering the time constraint and computation resources.
The competition consists of three phases; public, feedback, and final. Each phase provides five image datasets of different domains; people, objects, medical, aerial, and hand-writing (see Figure \ref{fig:dataset}). 
As one can see in Table \ref{table:autocv-datasets}, each dataset has different properties in the number of class, amounts of samples, resolution, and channel, even for the same domain but distinct phases. 
Furthermore, some dataset has variable image shapes; hence, the submitted code has to preprocess those data to adjust the unfixed dimension. 
These properties are veiled during the competition period. Therefore, participants must prepare the  automated algorithm which can reveal the meta-information of a given dataset without human intervention during the evaluation process. 

An import aspect of the competition is \emph{anytime learning}. The principal metric of AutoDL challenge is \emph{area under learning curve (ALC)} which drives the participants to submit a ready-to-predict learning algorithm at any timestamp during the time budget. The maximum time limit is 20 mins for each dataset. 
The competition also restricts computational resource and memory. Participants can only use a single GPU (NVIDIA Tesla P-100) with 4 CPU cores and 26GB memory during the evaluation process.
These restrictions and the evaluation metric implicitly block the computationally heavy AutoML approaches, including neural architecture search and hyperparameter optimization based on the distributed system e.g., Ray \cite{ray}.
Therefore, AutoCLINT applies a different strategy based on transfer learning and autonomous training strategy, including code optimization and automated augmentation. 

\begin{table}[t!] \center
\smaller
\begin{tabular}{c c c c c }
\toprule
  & & & \multicolumn{1}{c}{\# of Sample} & \multicolumn{1}{c}{Shape}  \\
 Phase & Dataset & \# of Class & (Train, Test) & (row, col, ch)
\tabularnewline
\midrule
Public & P & 26 & (80095, 19905) & (var, var, 3) \\
Public & O & 100 & (48061, 11939) & (32, 32, 3) \\
Public & M & 7 & (8050, 1965) & (600, 450, 3) \\
Public & A & 11 & (634, 166) & (var, var, 3) \\
Public & H & 10 & (60000, 10000) & (28, 28, 1) \\
\midrule
Feedback & P & 15 & (4406, 1094) & (350, 350, 3) \\
Feedback & O & 257 & (24518, 6089) & (var, var, 3) \\
Feedback & M & 2 & (175917, 44108) & (96, 96, 3) \\
Feedback & A & 3 & (324000, 81000) & (28, 28, 4) \\
Feedback & H & 3 & (6979, 1719) & (var, var, 3) \\
\midrule
Final & P & 100 & (6077, 1514) & (var, var, 3) \\
Final & O & 200 & (80000, 20000) & (32, 32, 3) \\
Final & M & 7 & (4492, 1114) & (976, 976, 3) \\
Final & A & 45 & (25231, 6269) & (256, 256, 3) \\
Final & H & 3 & (27938, 6939) & (var, var, 3)
\tabularnewline
\bottomrule
\end{tabular}
\caption{Summary of image datasets in AutoCV competitions. Each phase has five domains (P: People, O: Objects, M: Medical, A: Aerial, H: Hand-writing) and individual domain has different shape each other.}
\label{table:autocv-datasets}
\end{table}

\subsection{Evaluation Metric}
\label{sec:metric}

AutoCV challenge demands those as mentioned earlier anytime learning by scoring participants with the performance as a function of time. 
Each time after the training, the test code is called, and the results are saved. 
Then, the scoring program evaluates the model with their timestamp.

 \begin{figure*}
     \centering
     \includegraphics[width=0.8\textwidth]{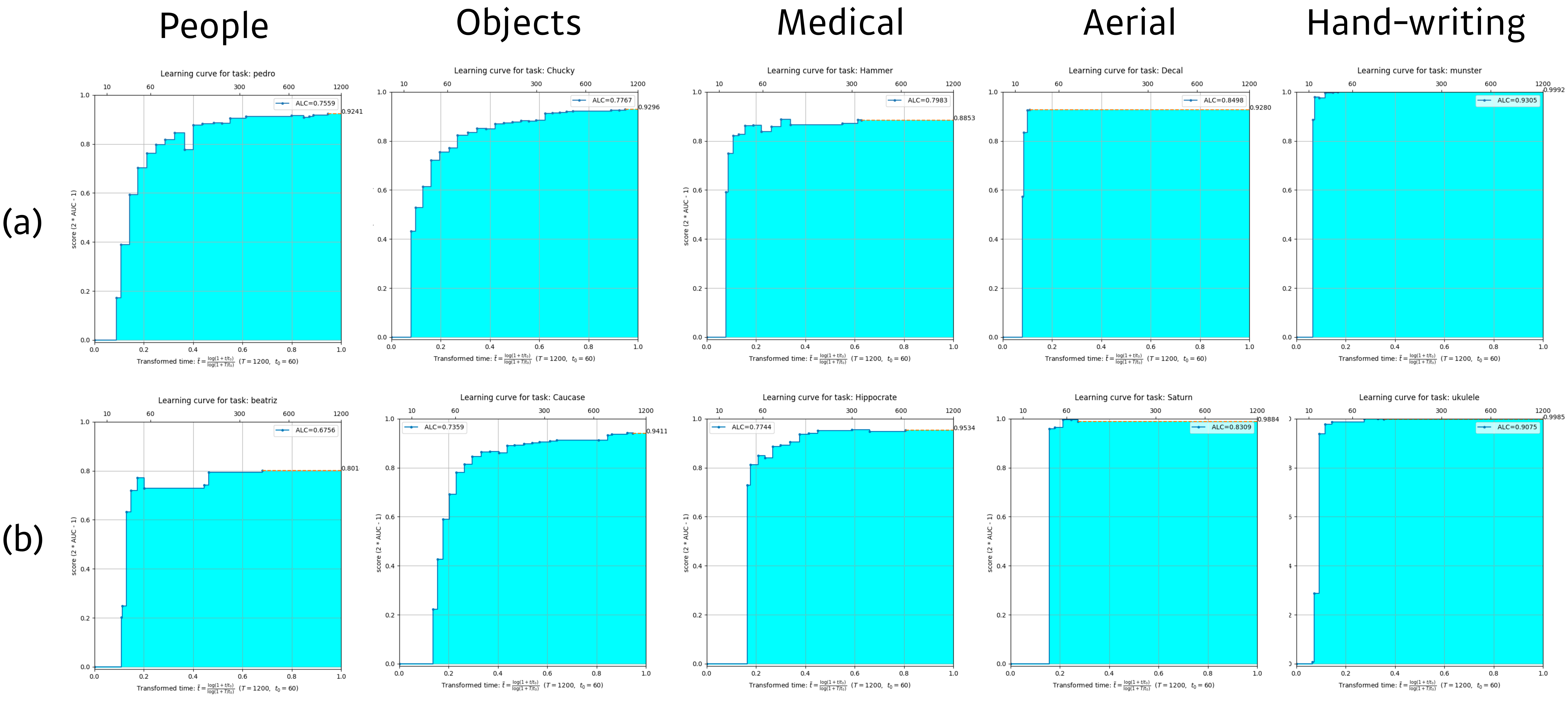}
     \caption{
     Learning curves of the proposed method on (a) public data and (b) feedback data. See Table 2 of the main paper for NAUC and ALC score.
     }
     \label{fig:ALC-competition}
 \end{figure*}

\paragraph{Normalized Area Under ROC Curve (NAUC)}

AutoCV competition treats each class as a separate binary classification problem. At each timestamp, the scoring program evaluates area under receiver operating characteristic curve (ROC AUC) for each prediction. Then, the \emph{normalized AUC (or Gini coefficient)} is computed by the following formula:
\begin{align}
    \label{eq:NAUC}
    \text{NAUC}=2\text{AUC} - 1.
\end{align}

\paragraph{Area under Learning Curve (ALC)}

ALC is the primary metric in AutoDL competitions. 
Due to \cite{liu:hal-02265053}, multi-class classification metrics are not being considered in AutoCV competition, i.e., each class is scored independently. 
AutoCV competition computed the ALC score for each dataset and used the overall ranking as the final score in the leaderboard. 
ALC is weighted in favor of the early convergence of the learning curve to achieve anytime learning. 

Let $\text{NAUC}(t)$ be the NAUC score at timestamp $t$, which is a step function with respect to time, then ALC is defined as follows:
\begin{align}
    \label{eq:ALC}
    \text{ALC}=\frac{1}{\log(1+T_b/t_{0})}\int_{0}^{T_b}\frac{\text{NAUC}(t)}{t+t_{0}}dt,
\end{align}
where $T_b$ is the time budget and $t_{0}$ is a reference time amount. In AutoCV competition, $T_b=1200$ and $t_{0}=60$ as a default. See Figure \ref{fig:ALC-competition} for the learning curve of the proposed method for each dataset in the public and feedback phase.

\section{AutoCLINT}
\label{sec:autoCLINT}

In this section, we provide the detail of components of the proposed method which is used in the AutoCV competition. At first, we introduce the autonomous training strategy in three-fold; (a) offline setting, (b) online data exploration, and (c) code optimization. 
As we mentioned above, NAS is not an appropriate strategy for fast training on arbitrarily given data since it takes a large amount of time for architecture search initially.
To get a higher score in the view of Area under Learning Curve (ALC) instead of Normalized Area Under ROC Curve (NAUC), we need an AutoML algorithm that can train the target network at once without wasting substantial time in the early architecture search phase and overfitting in the final phase.
In this case, \emph{Fast AutoAugment} \cite{lim2019fast} is a promising approach for training a given neural network on an arbitrary image domain.
We propose a light version of the Fast AutoAugment for limited computation resources and time constraints.


\subsection{Autonomous Training Strategy}
\label{sec:Auto-train}

\subsubsection{Offline Setting}

\paragraph{Architecture}
In both AutoCV and AutoCV2 competitions, we used ResNet18 for the network architecture. We also tried a bigger network, however it showed worse performances under a given time budget (see Figure \ref{fig:ablation-offline-arc}).
Before the main network blocks, our algorithm makes use of the meta information of given image data, including the spatial dimensions, the number of channels, and the number of classes. 
Then, the algorithm normalizes image data with 0.5-mean and 0.25-standard deviation. 
If the input channel is single, then we copy the channel into three channels. 
Otherwise, we prefix a $3 \times 3$ convolution layer to the main blocks.
Since the method for AutoCV is almost similar to the one for AutoCV2 competition, we only describe the former in this paper. 
The main difference is the split-up of the input data across time dimension and the multiplication of batch by the split data.

\paragraph{Pretraining and Softmax Annealing}

It is empirically known that ImageNet pretraining accelerates the convergence of models on the diverse image data \cite{kornblith2019better}.
Even though the pretrained model and randomly initialized model both achieve a similar final performance, the convergence speed is critical in the competition since the ALC score is sensitive to the early convergence due to the time-weighted integral in ALC (see (2)).
Hence, we employ the pretrained parameters of ResNet18 from \cmtt{torch.hub} except for the final fully connected layer which outputs the logit vector.
We initialize the fully connected layer by Xavier-normalization \cite{glorot2010understanding}.
Then we divide its output, the logit vector $\mathbf{z}=(\mathbf{z}_{1},\ldots,\mathbf{z}_{k})$, by $\tau$:
\begin{align}
    \label{eq:annealing}
    y_{i} = \frac{\exp(\mathbf{z}_{i}/\tau)}{\sum_{j=1}^{k}\exp(\mathbf{z}_{j}/\tau)}.
\end{align}
Softmax annealing with higher $\tau$ \eqref{eq:annealing} makes $y=(y_{1},\ldots,y_{k})$ uniform-like vector and marginally helps the training of the network in the view of ALC. We found out $\tau=8$ is optimal in the competition through the performance on public data.

\paragraph{Scheduled Optimizer}

We apply the scheduled SGD optimizer and warm-up \cite{goyal2017accurate} during 5 epochs in training for each dataset.
The learning rate decreases by $\frac{1}{10}$ if the train loss plateaus during 10 epochs. 
After the NAUC score is above 0.99, we search augmentation policies by Fast AutoAugment (see Section \ref{sec:fast-autoaugment}), restore the learning rate to the initial one (without warm-up), and retrain the network with the augmented train dataset.

\subsubsection{Online Setting}

\paragraph{Data Exploration}

Due to the variability in image size (see Table 1), the input tensor size of the network must be automatically adapted for each dataset to allow for adequate aggregation of spatial information and to keep the aspect ratio of the original image. 
The proposed method automatically adapt these parameters to the median size of each dataset, so that the network effectively trains on entire datasets. 
Due to time constraints, we do not increase the input tensor volume (without channels) beyond $64\times 64$. 
If the median shape of the dataset is smaller than $64\times 64$, then we use the median shape as the original input.

\paragraph{Code Optimization}

To fully utilize the computation resources, we optimize both preprocessing and training codes.
First, we generate transformed image data on CPU while GPU is utilized for training. 
We also implement multiple processing to parallelize the preparation of training images on multiple CPU cores. 
The validation and test data are transformed initially and reused on GPU memory since they are unchanged until the end of the training. 
Finally, CPU memory caches are operated to reduce the wasting time in disk I/O and common preprocessing, such as resizing.

\subsection{Automated Augment Strategy}
\label{sec:fast-autoaugment}

Data augmentation is beneficial to improve the generalization ability of deep learning networks. 
Recently, AutoAugment \cite{cubuk2018autoaugment} proposes the search space for automated augmentation and demonstrates that the obtained augmentation policies by reinforcement learning significantly improve the performances on the image classification tasks.
However, AutoAugment requires impractical time for search since it needs to train child network repeatedly.
Recently, Fast AutoAugment \cite{lim2019fast} overcomes this problem motivated from Bayesian hyperparameter optimization without the training of a given network twice in the search phase. 
Fast AutoAugment learns augmentation policies using a more efficient search strategy based on density matching. 
Ideally, Fast AutoAugment should be performed automatically, allowing the training data to adapt to test data.
AutoCLINT modifies the search space and implements a light version of Fast AutoAugment algorithm to surmount the restricted computational resources.

\paragraph{AutoAugment Strategy in AutoCLINT}
\label{sec:faa-autoclint}

Similar to Fast AutoAugment, the proposed method searches the augmentation policies that match the density of train data with the density of augmented valid data. 
We deviate from the original version in that we replace 5-fold with a single-fold search and use the random search (within a subset of searched policy in AutoAugment \cite{cubuk2018autoaugment}) instead of the Bayesian optimization on the original search space.

Let us explain the detail of search algorithm. For a given classification model $\mathcal{M}(\cdot|\theta)$ that is parameterized by $\theta$, an expected metric on dataset $D$ is denoted by $\mathcal{R}(\theta|D)$. We use the NAUC score for $\mathcal{R}$. Let $\mathcal{T}$ denote an augmentation policy which consists of sub-policies and each sub-policy has two consecutive image transformations. Every augmentation policy $\mathcal{T}$ outputs randomly transformed images since each transformation has the different calling probability (see \cite{cubuk2018autoaugment, lim2019fast} for the detail of search space in AutoAugment). We write $\mathcal{T}(D)$ for the augmented images of $D$ through policy $\mathcal{T}$.

\begin{table*}[t!] \center
\small
\begin{tabular}{c | c c | c c c c c }
\toprule
 & Rank & Time & People & Objects & Medical & Aerial & Hand-writing
\tabularnewline
\midrule
AutoCLINT (1st) & 4.0 & 46m 25s & \textbf{0.6756 (1)} & \textbf{0.7359 (1)} & \textbf{0.7744 (1)} & 0.8309 (16) & \textbf{0.9075 (1)} \\
Team A (2nd) & 6.4 & 57m 18s & 0.5815 (13) & 0.4918 (5) & 0.6682 (2) & \textbf{0.8675 (2)} & 0.8344 (10)  \\
Team B (3rd) & 9.2 & 56m 33s & 0.5692 (15) & 0.5574 (4) & 0.6387(5) & 0.8497 (10) & 0.8177 (12)
\tabularnewline
\bottomrule
\end{tabular}
\caption{The comparison of ALC  and rank on each data at the feedback phase of AutoCV. We only list the result of winning teams at the final phase. The proposed method won the top ALC score (except Aerial data) and recorded the fastest computation time in the competition.}
\label{table:autocv-result}
\end{table*}

\begin{table*}[t!] \center
\small
\begin{tabular}{c | c c | c c c c c }
\toprule
 & Rank & Time & People & Hand-writing & Action1 & Action2 & Action3
\tabularnewline
\midrule
AutoCLINT (1st) & 5.2 & 1h 12m 15s & 0.6277 (8) & 0.9048 (5) & 0.4076 (8) & \textbf{0.4640 (2)} & 0.2091 (3) \\
Team C (2nd) & 6.2 & 1h 11m 28s & 0.6231 (9) & 0.8406 (11) & 0.4527 (4) & 0.3688 (6) & \textbf{0.2363 (1)} \\
Team D (3rd) & 6.2 & 1h 05m 13s & \textbf{0.6835 (2)} & \textbf{0.9115 (3)} & \textbf{0.4658 (2)} & -0.0417 (16) & 0.1627 (8)

\tabularnewline
\bottomrule
\end{tabular}
\caption{The comparison of ALC  and rank on each data at the final phase of AutoCV2.}
\label{table:autocv-result2}
\end{table*}

\begin{table*}[t!] \center
\small
\begin{tabular}{c c | c c | c c | c c | c c }
\toprule
  & & \multicolumn{2}{c|}{Linear} & \multicolumn{2}{c|}{ResNet50-V2} & \multicolumn{2}{c|}{Inception-V3} & \multicolumn{2}{c}{AutoCLINT} \\
 Phase & Dataset & NAUC & ALC & NAUC & ALC & NAUC & ALC & NAUC & ALC 
\tabularnewline
\midrule
Public & People & 0.2863 & 0.1733 & 0.8009 & 0.2115 & 0.5867 & 0.1805 & 0.9214 & \textbf{0.7366} \\
Public & Objects & 0.2331 & 0.1643 & 0.7877 & 0.1914 & 0.9270 & 0.3289 & 0.9353 & \textbf{0.7835} \\
Public & Medical & 0.1922 & 0.1596 & 0.1173 & 0.0238 & 0.7986 & 0.4742 & 0.9142 & \textbf{0.8286} \\
Public & Aerial & 0.0982 & 0.0893 & 0.5833 & 0.2085 & 0.8861 & 0.5712 & 0.9347 & \textbf{0.8353} \\
Public & Hand-writing & 0.9628 & 0.8223 & 0.9999 & 0.5408 & 0.9950 & 0.5883 & 0.9977 & \textbf{0.9440} \\
\midrule
Feedback & People & 0.2129 & 0.1829 & 0.5675 & 0.1350 & 0.6621 & 0.3212 & 0.8014 & \textbf{0.6756} \\
Feedback & Objects & 0.1249 & 0.0683 & 0.7801 & 0.1910 & 0.9897 & 0.3367 & 0.9411 & \textbf{0.7359} \\
Feedback & Medical & 0.3743 & 0.1726 & 0.8314 & 0.2319 & 0.8452 & 0.4571 & 0.9534 & \textbf{0.7744} \\
Feedback & Aerial & 0.9003 & 0.3507 & 0.9987 & 0.2860 & 0.9665 & 0.3621 & 0.9884 & \textbf{0.8309} \\
Feedback & Hand-writing & 0.3003 & 0.2747 & 0.9986 & 0.5093 & 0.4189 & 0.1276 & 0.9985 & \textbf{0.9075}
\tabularnewline
\bottomrule
\end{tabular}
\caption{The comparison between proposed method and baseline methods \cite{liu:hal-02265053} on public data and feedback data of AutoCV.}  
\label{table:autocv-public-feedback}
\vspace{-0.2cm}
\end{table*}

\SetKwInOut{Input}{Input}
\SetKwInOut{Output}{Output}

\begin{algorithm}[t]
\Input{$\left(\mathcal{S}, D_{\text{train}}, D_{\text{valid}}, C, T, N \right)$}

    $\mathcal{B} \leftarrow \emptyset$ \\
    Train $\theta$ on $D_{\text{train}}$ \\
    Evaluate $\mathcal{R}(\theta|D_{\text{valid}})$ 
    
    \For{$t = 1,\ldots,T$}
    {
    $\mathcal{T}_{1:C} \leftarrow \text{RandomChoice}(\mathcal{S}, C)$ \\
    Evaluate $\mathcal{R}(\theta|\mathcal{T}_{1:C}(D_{\text{valid}}))$ \\
    \If{$\mathcal{R}(\theta|\mathcal{T}_{1:C}(D_{\text{valid}})) > \mathcal{R}(\theta|D_{\text{valid}})$}{
        Append $\mathcal{T}_{1:C}$ to $\mathcal{B}$ \\
    }
    
    }
Select Top-$N$ policies from $\mathcal{B}$
\caption{Fast AutoAugment in AutoCLINT}
\label{alg:fast-autoaugment}
\end{algorithm}

Assuming $\theta$ is trained on $D_{\text{train}}$ sufficiently, we randomly choose $C$ sub-policies from the collection of augmentation policies $\mathcal{S}$, which is already found by AutoAugment search algorithm on CIFAR-10. 
Let $\mathcal{T}_{1:C}$ denote a policy composed of sampled sub-policies.
Then, we evaluate $\mathcal{R}(\theta | \mathcal{T}_{1:C}(D_{\text{valid}}))$ and append $\mathcal{T}_{1:C}$ to a list $\mathcal{B}$ unless the score is lower than $\mathcal{R}(\theta|D_{\text{valid}})$. 
We repeat the above process $T$ times. 
After the procedure, we select top-$N$ policies from the list of searched policies $\mathcal{B}$ according to the score. 
In the competition, we set $C=3, T=100, N=5$. See Algorithm \ref{alg:fast-autoaugment} for summary. 

Instead of selecting top policies from the list $\mathcal{B}$, one can merely apply augmentation policies at random in Algorithm \ref{alg:fast-autoaugment} without evaluating the augmentation policies at the validation dataset. This strategy can reduce search time more than the proposed algorithm. In this case, however, both ALC and NAUC are empirically worse than no augmentation strategy. We empirically show a comparison between the proposed algorithm and random augmentation in Section 4.

\subsection{Competition Results}

As a result, AutoCLINT won the top place at both AutoCV and AutoCV2 competitions in the final leaderboard\footnote{https://autodl.lri.fr/competitions/3\#home}. The proposed method showed the highest average rank. (see Table \ref{table:autocv-result}, \ref{table:autocv-result2}). 
Notably, the proposed method records the fastest training and outperforms the other approaches from the perspective of ALC on every dataset except aerial data at AutoCV competition. Here, the other participants achieve similar ALC scores on aerial data. 
Hence, the proposed method demonstrates its efficiency in the competitions.
Furthermore, according to Table \ref{table:autocv-public-feedback}, AutoCLINT recorded a superior NAUC score in every image domain. 
Therefore, it is proven that the proposed method is anytime learnable and has excellent generalization performance for various image datasets.

\section{Empirical Studies}
\label{sec:ablation-studies}

In this section, we validate the effect of each component of AutoCLINT based on empirical results. We first present comparison between the proposed method and baseline models. Then, we analyze the effect of offline setting; (a) architectures, (b) pretraining, and (c) softmax annealing. Ablation results are provided among different offline settings. Also, we examine the proposed automated augmentation strategies compared to no augmentation and random augmentation. All scores are measured on public datasets since the other datasets are not open publicly. 

\subsection{Comparison to Baseline}
\label{sec:baseline}

AutoCV white paper \cite{liu:hal-02265053} provides the baseline methods (linear, ResNet50-v2, pretrained Inception) and evaluates metrics on both public and feedback data. 
Likewise, we evaluate the proposed method on the same datasets. 

AutoCLINT outperforms the baseline methods in ALC. 
See Table 3 in the main paper for the comparison between the proposed method and baseline methods on each dataset.
Contrary to the public phase, AutoCLINT shows decreased ALC score at the feedback phase. However, the proposed method shows outstanding performance in NAUC score at the feedback phase (except people), while baseline models show deteriorated scores in NAUC. 
This result demonstrates that AutoCLINT is transferable for image classification tasks.

\subsection{Effect of Offline Setting}

\begin{figure}
     \centering
     \includegraphics[width=0.4\textwidth]{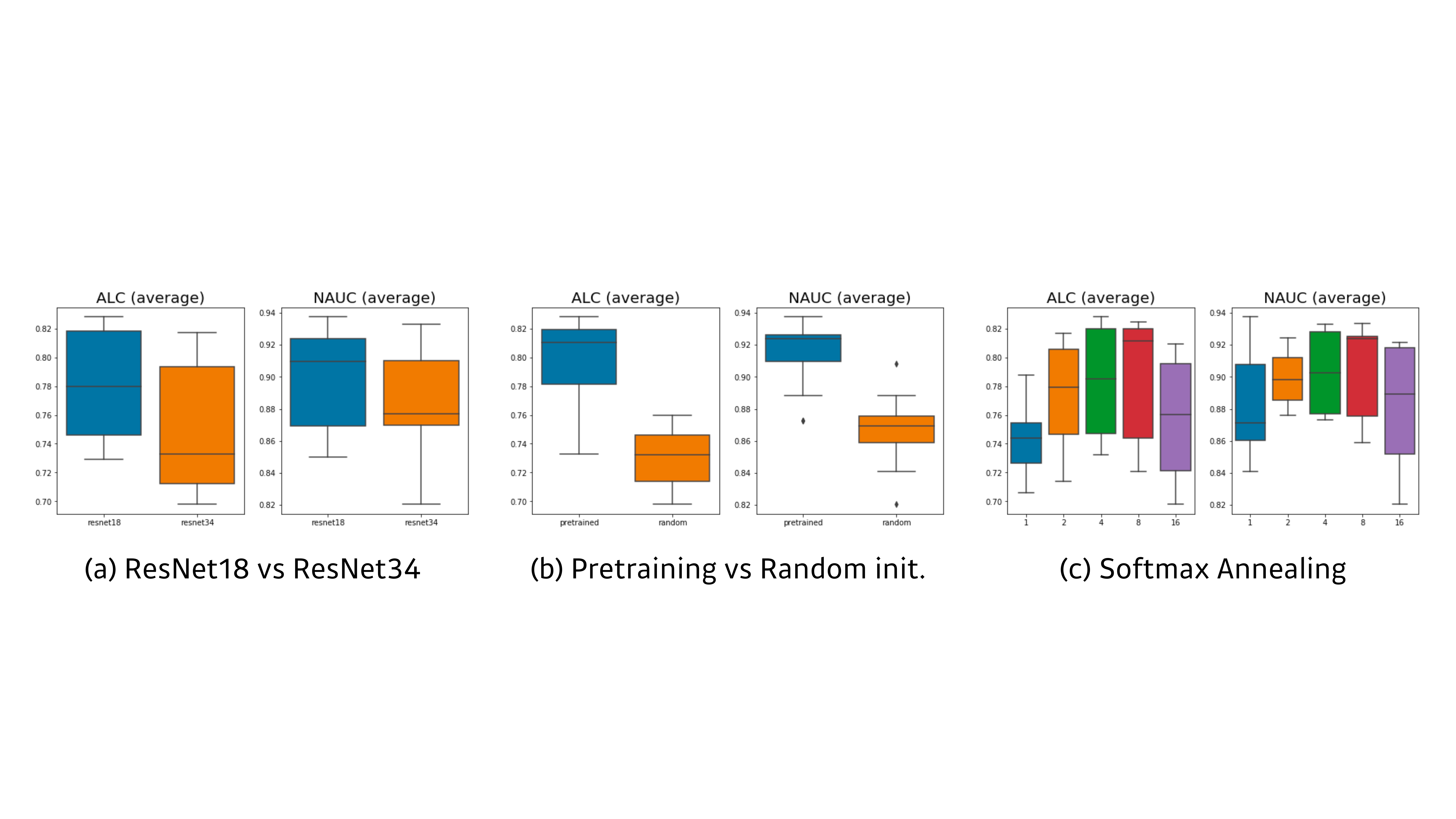}
     \caption{
     Boxplots of average ALC and average NAUC between ResNet18 and ResNet34 provided the other settings are fixed.
     }
     \label{fig:ablation-offline-arc}
 \end{figure}

\begin{figure}
     \centering
     \includegraphics[width=0.4\textwidth]{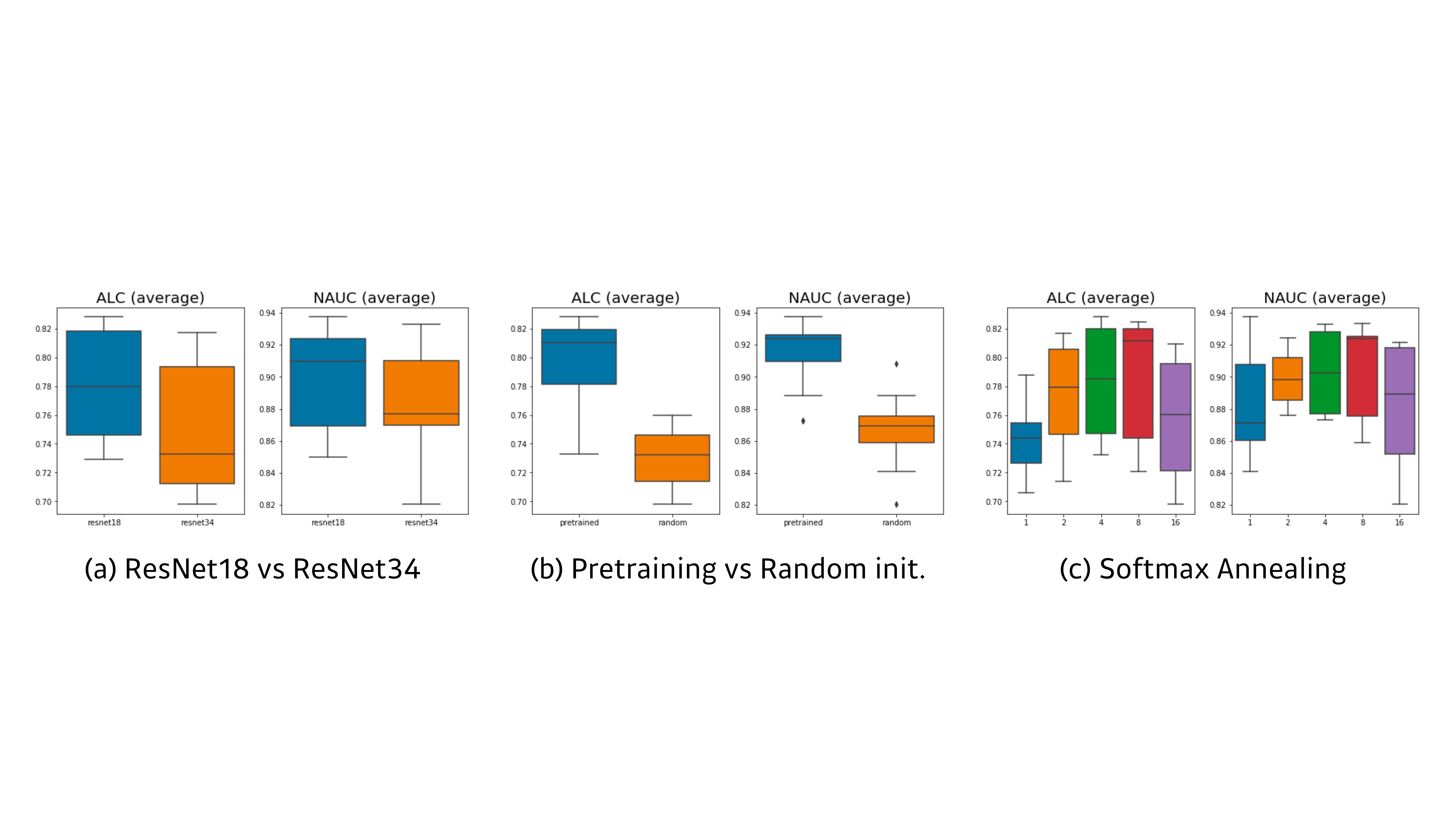}
     \caption{
     Boxplots of average ALC and average NAUC between pretrained network and randomly initialized network. 
     }
     \label{fig:ablation-offline-init}
 \end{figure}

\begin{figure}
     \centering
     \includegraphics[width=0.4\textwidth]{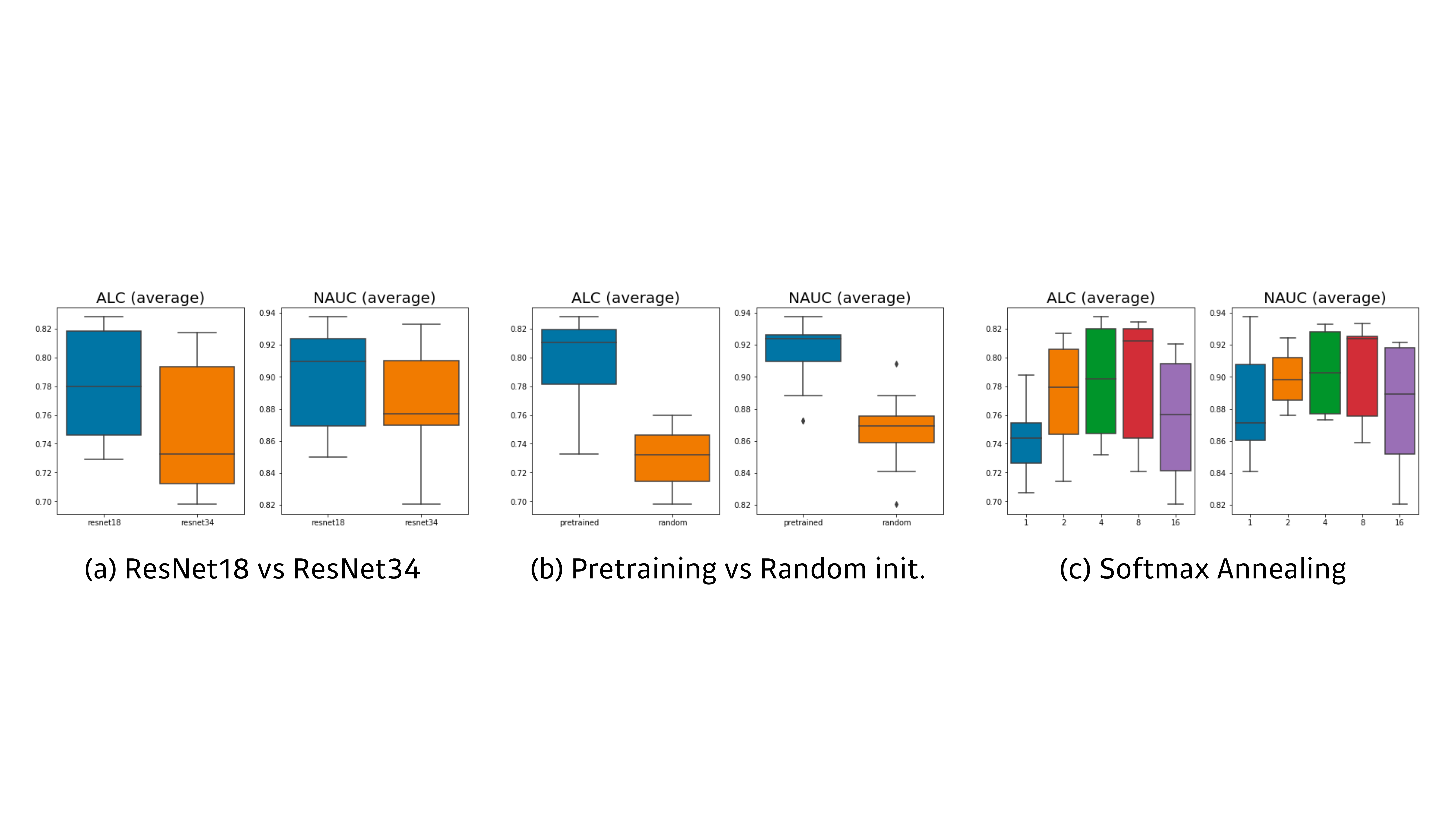}
     \caption{
     Boxplots of average ALC and average NAUC among $\tau=1,2,4,8,16$.
     }
     \label{fig:ablation-offline-tau}
 \end{figure}

We measure metrics by changing the components of AutoCLINT in three ways: (a) ResNet18 vs. ResNet34, (b) Pretraining vs. Random Initialization, and (c) Softmax Annealing. 
Figure \ref{fig:ablation-offline-arc}-\ref{fig:ablation-offline-tau} shows the statistical difference between variations of the components. In summary, every component is beneficial in AutoCLINT; especially, pretraining is the most necessary component in the proposed method.

\paragraph{ResNet18 vs. ResNet34}

Figure \ref{fig:ablation-offline-arc} shows that ResNet18 is slightly better than ResNet34 in both ALC and NAUC. 
Since ResNet18 is computationally lighter than ResNet34, the ALC score of ResNet18 is higher than the ALC score of ResNet34. 
The NAUC score of ResNet34 is slightly lower than that of ResNet18 because the training time and computation resources are insufficient to train ResNet34.
It is expected that if the more time budget is given, then a bigger model may show better performance. 
However, because only 20 minutes are given for each domain, we fix ResNet18 as architecture in the competition.

\paragraph{Pretraining vs. Random Initialization}

Pretraining contributes more than the other offline setting components in the proposed method. Figure \ref{fig:ablation-offline-init} shows that the ImageNet-pretrained ResNet18 overwhelms the randomly initialized networks in both metrics. 
Indeed, pretrained networks report a higher NAUC score on each dataset.
This is an expected result since the pretraining generally helps the filters of convolution networks to learn how to discriminate features in image data. 
These results fairly coincide with the analysis of \cite{kornblith2019better}.

 \begin{figure}
     \centering
     \includegraphics[width=0.4\textwidth]{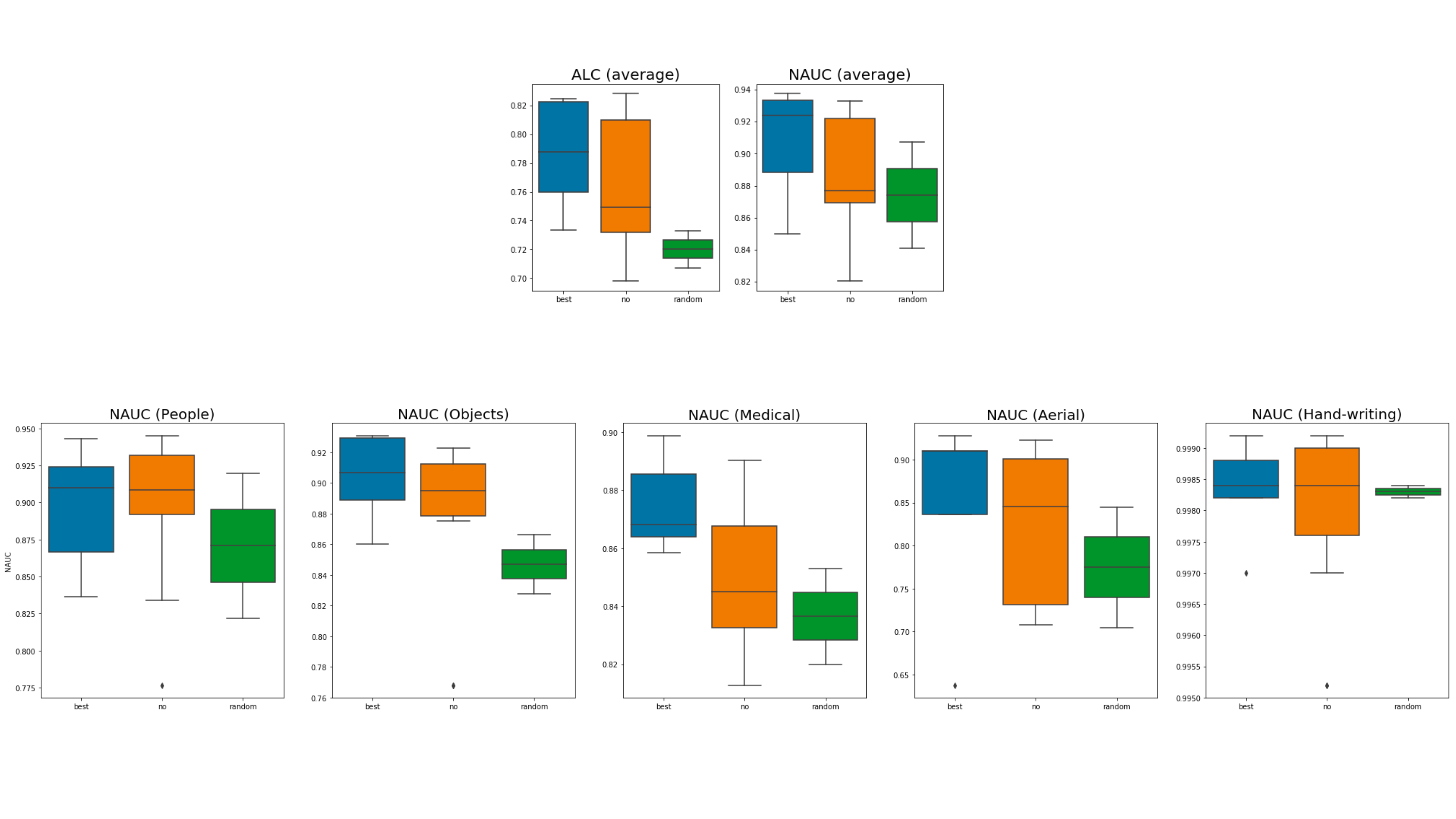}
     \caption{
     Boxplots of average ALC and average NAUC among different augmentation strategies.
     }
     \label{fig:ablation-faa}
 \end{figure}

 \begin{figure*}
     \centering
     \includegraphics[width=0.9\textwidth]{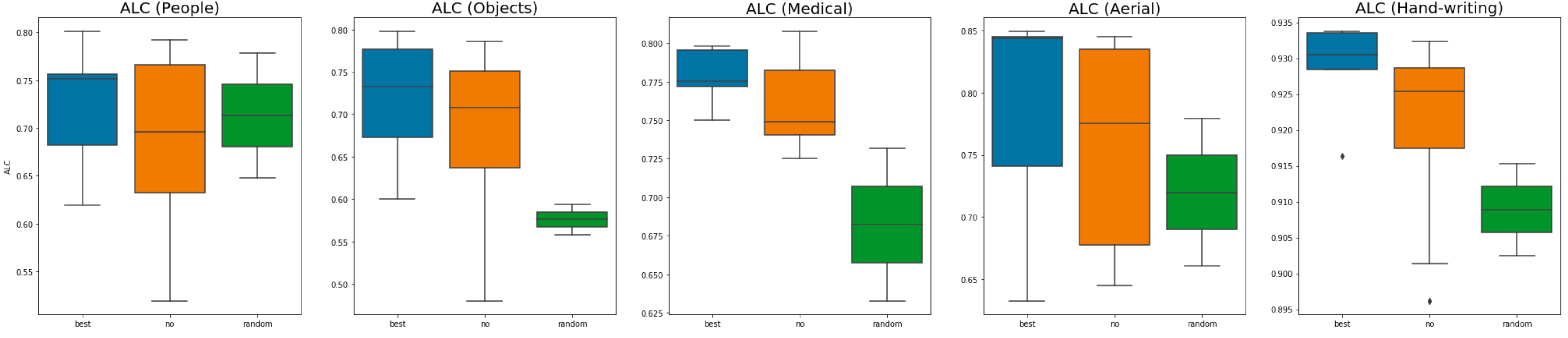}
     \caption{
     Boxplots of ALC among different augmentation strategies on each dataset.
     }
     \label{fig:ablation-faa-data-alc}
 \end{figure*}

 \begin{figure*}
     \centering
     \includegraphics[width=0.9\textwidth]{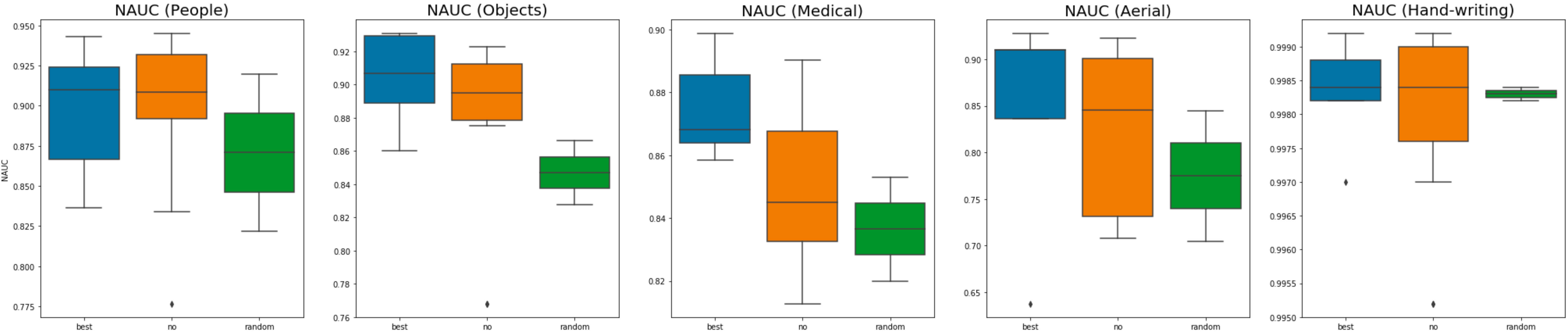}
     \caption{
     Boxplots of NAUC among different augmentation strategies on each dataset.
     }
     \label{fig:ablation-faa-data}
 \end{figure*}

\paragraph{Softmax Annealing}

Softmax annealing with higher $\tau$ makes the output vector as a uniform-like one. We observe that it marginally improves the performance. Figure \ref{fig:ablation-offline-tau} shows that $\tau=2, 4, 8$ are better than no annealing ($\tau=1$) in ALC. 
However, the effect of annealing decreases when $\tau=16$ and it shows the lower result than no annealing in the worst case. 
Hence we fix $\tau=8$ during the competition.

\subsection{Effect of Automated Augmentation}

The technical components introduced in Section 2 of the main paper are practical but well-known heuristics in deep learning communities. 
Therefore, the proposed automated augmentation is a unique feature in AutoCLINT compared to the other approaches in the competition. In this section, we provide ablation studies regarding the effect of the proposed augmentation strategies.

Figure \ref{fig:ablation-faa-data-alc} supports the efficiency of Algorithm 1 in the view of anytime learning. ALC score on each domain of the proposed method is higher than the other policies.
Notably, objects, medical, and aerial domains get better NAUC scores than no augmentation (see Figure \ref{fig:ablation-faa-data}). 
According to Table \ref{table:autocv-datasets}, these image domains have less number of samples than the other domains (people and hand-writing) in the public phase. 
Thus, the proposed algorithm can help to find augmentation policies to make the target network learns better. Figure \ref{fig:ablation-faa} shows that the proposed augmentation search algorithm significantly increases metrics on average. 

Note that random augmentation decreases the performance of network rather than increase. 
These results indicate that Algorithm 1 is efficient in finding augmentation policies automatically for any image domain.

\section{Related Works}

\paragraph{Automated Machine Learning} 

AutoML is a subfield of meta-learning and continuously draws attention from industry and academia even before the great success of deep learning.
We refer to \cite{zoller2019survey} and references therein for the survey of the AutoML framework on classical machine learnings. 
The runtime of hyperparameter optimization is reduced from several hours to mere minutes. \cite{hutter2019automated}.

Due to the dramatic success of deep learning in the visual domain, machine learning researchers have focused on neural architecture search algorithms that can find the optimal neural network for arbitrarily given data. NAS has a significant overlap with hyperparameter optimization and meta learning. We refer to \cite{elsken2018neural} and references therein for the detailed research flow of NAS. 
Recent researches show that reinforcement learning (RL) works well in the NAS framework \cite{zoph2016neural, pham2018efficient}; however, such RL based methods are not applicable to AutoCV competition since they mostly require child network training or proxy tasks, which are obstacles to anytime learning.
ProxylessNAS \cite{cai2018proxylessnas} proposes a gradient-based NAS algorithm that can directly find the neural networks for target tasks and hardware platforms. 

NAS is a promising technique in AutoML, but it is not an efficient strategy in the competition, which restricts the time budget within 20 minutes for each domain. Hence AutoCLINT uses a transfer learning strategy with automated augmentation.

\paragraph{Automated Data Augmentation} 

Data augmentation is a necessary procedure for improving the generalization ability of deep neural network models.
A carefully designed set of augmentations in training improves the performance of a network significantly. 
Adequately designed data augmentations in training improve the performance of a model significantly. 
However, such planning has relied heavily on domain experts. 
Automated data augmentation is a subfield of AutoML research and recently draw attention from machine learning researchers, especially for computer vision tasks since they report the state-of-the-art performance. AutoAugment \cite{cubuk2018autoaugment} proposes the search space of augmentation policies for image data and the search algorithm based on reinforcement learning for the classification task and detection task \cite{zoph2019learning}. 
However, AutoAugment requires thousands of GPU hours even in a reduced setting. 
To overcome the time inefficiency of AutoAugment, PBA \cite{ho2019pba} and Fast AutoAugment \cite{lim2019fast} propose different search algorithms based on hyperparameter optimization. On the other hand, Randaugment \cite{cubuk2019randaugment} exceeds the predictive performance of other augmentation methods with a significantly reduced search space. 

AutoCLINT uses augmentation policies from AutoAugment and implements an automated data augmentation algorithm motivated from Fast AutoAugment to reduce the search cost under the restricted time resource. As a result, the proposed method outperforms the other approaches under the time constraint.

\section{Conclusion}

In this paper, we provide technical details of AutoCLINT, the winning method for AutoCV competitions. AutoCLINT relies on autonomous training strategy and automated data augmentation, and it demonstrates its learning efficiency. We expect the proposed method is widely applicable to machine learning problems under limited time and computational resources.

\bibliographystyle{ieee_fullname}
\bibliography{bib_autoclint}

\begin{thebibliography}{10}\itemsep=-1pt

\bibitem{bergstra2011algorithms}
James~S Bergstra, R{\'e}mi Bardenet, Yoshua Bengio, and Bal{\'a}zs K{\'e}gl.
\newblock Algorithms for hyper-parameter optimization.
\newblock In {\em Advances in Neural Information Processing Systems}, pages
  2546--2554, 2011.

\bibitem{cai2018proxylessnas}
Han Cai, Ligeng Zhu, and Song Han.
\newblock {ProxylessNAS}: Direct neural architecture search on target task and
  hardware.
\newblock In {\em 7th International Conference on Learning Representations,
  {ICLR} 2019, New Orleans, LA, USA, May 6-9, 2019}, 2019.

\bibitem{cubuk2018autoaugment}
Ekin~D Cubuk, Barret Zoph, Dandelion Mane, Vijay Vasudevan, and Quoc~V Le.
\newblock Autoaugment: Learning augmentation strategies from data.
\newblock In {\em Proceedings of the IEEE conference on computer vision and
  pattern recognition}, 2019.

\bibitem{cubuk2019randaugment}
Ekin~D Cubuk, Barret Zoph, Jonathon Shlens, and Quoc~V Le.
\newblock Randaugment: Practical data augmentation with no separate search.
\newblock {\em arXiv preprint arXiv:1909.13719}, 2019.

\bibitem{elsken2018neural}
Thomas Elsken, Jan~Hendrik Metzen, and Frank Hutter.
\newblock Neural architecture search: A survey.
\newblock {\em arXiv preprint arXiv:1808.05377}, 2018.

\bibitem{glorot2010understanding}
Xavier Glorot and Yoshua Bengio.
\newblock Understanding the difficulty of training deep feedforward neural
  networks.
\newblock In {\em Proceedings of the thirteenth international conference on
  artificial intelligence and statistics}, pages 249--256, 2010.

\bibitem{goyal2017accurate}
Priya Goyal, Piotr Doll{\'a}r, Ross Girshick, Pieter Noordhuis, Lukasz
  Wesolowski, Aapo Kyrola, Andrew Tulloch, Yangqing Jia, and Kaiming He.
\newblock Accurate, large minibatch sgd: Training imagenet in 1 hour.
\newblock {\em arXiv preprint arXiv:1706.02677}, 2017.

\bibitem{ho2019pba}
Daniel Ho, Eric Liang, Ion Stoica, Pieter Abbeel, and Xi Chen.
\newblock Population based augmentation: Efficient learning of augmentation
  policy schedules.
\newblock In {\em ICML}, 2019.

\bibitem{hutter2019automated}
Frank Hutter, Lars Kotthoff, and Joaquin Vanschoren.
\newblock Automated machine learning-methods, systems, challenges, 2019.

\bibitem{kim2018scalable}
Sungwoong Kim, Ildoo Kim, Sungbin Lim, Woonhyuk Baek, Chiheon Kim, Hyungjoo
  Cho, Boogeon Yoon, and Taesup Kim.
\newblock Scalable neural architecture search for 3d medical image
  segmentation.
\newblock In {\em Medical Image Computing and Computer Assisted Intervention --
  MICCAI 2019}, pages 220--228. Springer International Publishing, 2019.

\bibitem{kornblith2019better}
Simon Kornblith, Jonathon Shlens, and Quoc~V Le.
\newblock Do better imagenet models transfer better?
\newblock In {\em Proceedings of the IEEE Conference on Computer Vision and
  Pattern Recognition}, pages 2661--2671, 2019.

\bibitem{lim2019fast}
Sungbin Lim, Ildoo Kim, Taesup Kim, Chiheon Kim, and Sungwoong Kim.
\newblock Fast autoaugment.
\newblock In {\em Advances in Neural Information Processing Systems (NeurIPS)},
  2019.

\bibitem{liu2018darts}
Hanxiao Liu, Karen Simonyan, and Yiming Yang.
\newblock Darts: Differentiable architecture search.
\newblock {\em arXiv preprint arXiv:1806.09055}, 2018.

\bibitem{liu:hal-02265053}
Zhengying Liu, Isabelle Guyon, Julio~Jacques Junior, Meysam Madadi, Sergio
  Escalera, Adrien Pavao, Hugo~Jair Escalante, Wei-Wei Tu, Zhen Xu, and
  Sebastien Treguer.
\newblock {AutoCV Challenge Design and Baseline Results}.
\newblock In {\em {CAp 2019 - Conf{\'e}rence sur l'Apprentissage Automatique}},
  Toulouse, France, July 2019.

\bibitem{ray}
Philipp Moritz, Robert Nishihara, Stephanie Wang, Alexey Tumanov, Richard Liaw,
  Eric Liang, Melih Elibol, Zongheng Yang, William Paul, Michael~I Jordan,
  et~al.
\newblock Ray: A distributed framework for emerging $\{$AI$\}$ applications.
\newblock In {\em 13th $\{$USENIX$\}$ Symposium on Operating Systems Design and
  Implementation ($\{$OSDI$\}$ 18)}, pages 561--577, 2018.

\bibitem{pham2018efficient}
Hieu Pham, Melody Guan, Barret Zoph, Quoc Le, and Jeff Dean.
\newblock Efficient neural architecture search via parameter sharing.
\newblock In {\em International Conference on Machine Learning}, pages
  4092--4101, 2018.

\bibitem{zoller2019survey}
Marc-Andr{\'e} Z{\"o}ller and Marco~F Huber.
\newblock Survey on automated machine learning.
\newblock {\em arXiv preprint arXiv:1904.12054}, 2019.

\bibitem{zoph2019learning}
Barret Zoph, Ekin~D Cubuk, Golnaz Ghiasi, Tsung-Yi Lin, Jonathon Shlens, and
  Quoc~V Le.
\newblock Learning data augmentation strategies for object detection.
\newblock {\em arXiv preprint arXiv:1906.11172}, 2019.

\bibitem{zoph2016neural}
Barret Zoph and Quoc~V. Le.
\newblock Neural architecture search with reinforcement learning.
\newblock In {\em International Conference on Learning Representations}, 2017.

\bibitem{zoph2018learning}
Barret Zoph, Vijay Vasudevan, Jonathon Shlens, and Quoc~V Le.
\newblock Learning transferable architectures for scalable image recognition.
\newblock In {\em Proceedings of the IEEE conference on computer vision and
  pattern recognition}, pages 8697--8710, 2018.

\end{thebibliography}

\end{document}